\pdfoutput=1

\documentclass[11pt]{article}

\usepackage{ACL2023}

\usepackage{times}
\usepackage{latexsym}

\usepackage[T1]{fontenc}

\usepackage[utf8]{inputenc}

\usepackage{microtype}

\usepackage{inconsolata}

\usepackage{booktabs}
\usepackage{amssymb}
\usepackage{svg}
\usepackage{mathtools}
\usepackage{algorithm}
\usepackage[noend]{algpseudocode}

\title{Disambiguating Numeral Sequences to\\Decipher Ancient Accounting Corpora}

\author{
    \begin{tabular}{c}
    Logan Born$^1$\\
    {\tt\small loborn@sfu.ca}
    \end{tabular}
    \begin{tabular}{c}
    M. Willis Monroe$^2$\\
    {\tt\small willis.monroe@ubc.ca}
    \end{tabular}
    \begin{tabular}{c}
    Kathryn Kelley$^3$\\
    {\tt\small kathrynerin.kelley@unibo.it}
    \end{tabular}
    \begin{tabular}{c}
    Anoop Sarkar$^1$\\
    {\tt\small anoop@cs.sfu.ca}
    \end{tabular}\\[12pt]
    \begin{tabular}{c}
         $^1$Simon Fraser University\\
         School of Computing Science
    \end{tabular} $\quad$
    \begin{tabular}{c}
         $^2$University of British Columbia\\
         Department of Philosophy
    \end{tabular}\\[12pt]
    \begin{tabular}{c}
         $^3$Universit\`a di Bologna\\
         Dipartimento di Filologia Classica e Italianistica
    \end{tabular} \\[12pt]
}

\begin{document}

\newcommand{\NOne}[0]{\raisebox{1.0em}{\includegraphics[width=0.2in,angle=-90]{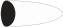}} }
\newcommand{\NFourteen}[0]{\raisebox{0.7em}{\includegraphics[width=0.1in,angle=-90]{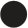}} }
\newcommand{\NThirtyFour}[0]{\raisebox{0.7em}{\includegraphics[width=0.25in,angle=-90]{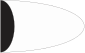}} }

\maketitle
\begin{abstract}
A \textit{numeration system} encodes abstract numeric quantities as concrete strings of written characters. 
The numeration systems used by modern scripts tend to be precise and unambiguous, but this was not so for the ancient and partially-deciphered proto-Elamite (PE) script, where written numerals can have up to four distinct readings depending on the system that is used to read them. 
We consider the task of disambiguating between these readings in order to determine the values of the numeric quantities recorded in this corpus.
We algorithmically extract a list of possible readings for each PE numeral notation, and contribute two disambiguation techniques based on structural properties of the original documents and classifiers learned with the bootstrapping algorithm.
We also contribute a test set for evaluating disambiguation techniques, as well as a novel approach to cautious rule selection for bootstrapped classifiers.
Our analysis confirms existing intuitions about this script and reveals previously-unknown correlations between tablet content and numeral magnitude.
This work is crucial to understanding and deciphering PE, as the corpus is heavily accounting-focused and contains many more numeric tokens than tokens of text.
\end{abstract}

\section{Introduction\footnote{\citealt{englund1996} contains a printing error in the table which reports the relative values of signs in the decimal system. An earlier version of this paper relied on the erroneous values from that publication. This is an updated version which corrects those mistakes.}}
\begin{figure}
    \centering
    \textbf{Sexagesimal}
    
    \raisebox{1em}{
        \includegraphics[width=0.3in,angle=-90]{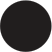}
    }
    $\xleftarrow{6}$
    \raisebox{1.1em}{
        \includegraphics[width=0.3in,angle=-90]{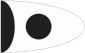}
    }
    $\xleftarrow{10}$
    \raisebox{1.1em}{
        \includegraphics[width=0.3in,angle=-90]{figures/signs/1N34.png}
    }
    $\xleftarrow{6}$
    \raisebox{0.7em}{
        \includegraphics[width=0.15in,angle=-90]{figures/signs/1N14.png}
    }
    $\xleftarrow{10}$
    \raisebox{1.1em}{
        \includegraphics[width=0.2in,angle=-90]{figures/signs/1N01.png}
    }
    $\xleftarrow{2}$
    \raisebox{0.7em}{
        \includegraphics[width=0.15in,angle=-90]{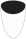}
    }\\[1em]
    
    \textbf{Decimal}
    
    \raisebox{1.2em}{
        \includegraphics[width=0.35in,angle=-90]{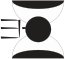}
    }
    $\xleftarrow{10}$
    \raisebox{1.1em}{
        \includegraphics[width=0.3in,angle=-90]{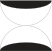}
    }
    $\xleftarrow{10}$
    \raisebox{1.0em}{
        \includegraphics[width=0.3in,angle=-90]{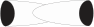}
    }
    $\xleftarrow{10}$
    \raisebox{0.7em}{
        \includegraphics[width=0.15in,angle=-90]{figures/signs/1N14.png}
    }
    $\xleftarrow{10}$
    \raisebox{0.7em}{
        \includegraphics[width=0.2in,angle=-90]{figures/signs/1N01.png}
    }\\[1em]
    
    \textbf{Bisexagesimal}
    
    \raisebox{1em}{
        \includegraphics[width=0.3in,angle=-90]{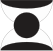}
    }
    $\xleftarrow{10}$
    \raisebox{1.1em}{
        \includegraphics[width=0.3in,angle=-90]{figures/signs/1N51.png}
    }
    $\xleftarrow{2}$
    \raisebox{1.1em}{
        \includegraphics[width=0.3in,angle=-90]{figures/signs/1N34.png}
    }
    $\xleftarrow{6}$
    \raisebox{0.7em}{
        \includegraphics[width=0.15in,angle=-90]{figures/signs/1N14.png}
    }
    $\xleftarrow{10}$
    \raisebox{0.7em}{
        \includegraphics[width=0.2in,angle=-90]{figures/signs/1N01.png}
    }\\[1em]
    
    \textbf{Capacity}
    
    \raisebox{1em}{
        \includegraphics[width=0.35in,angle=-90]{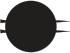}
    }
    $\xleftarrow{6}$
    \raisebox{1.1em}{
        \includegraphics[width=0.3in,angle=-90]{figures/signs/1N48.png}
    }
    $\xleftarrow{10}$
    \raisebox{1.1em}{
        \includegraphics[width=0.3in,angle=-90]{figures/signs/1N34.png}
    }
    $\xleftarrow{3}$
    \raisebox{1.1em}{
        \includegraphics[width=0.3in,angle=-90]{figures/signs/1N45.png}
    }
    $\xleftarrow{10}$
    \raisebox{0.7em}{
        \includegraphics[width=0.15in,angle=-90]{figures/signs/1N14.png}
    }
    $\xleftarrow{6}$
    \raisebox{0.7em}{
        \includegraphics[width=0.2in,angle=-90]{figures/signs/1N01.png}
    }
    $\xleftarrow{5}$
    \raisebox{1.0em}{
        \includegraphics[width=0.2in,angle=-90]{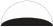}
    }
    $\xleftarrow{2}$
    \raisebox{0.7em}{
        \includegraphics[width=0.1in,angle=-90]{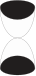}
    }
    $\xleftarrow{3}$
    \raisebox{0.8em}{
        \includegraphics[width=0.2in,angle=-90]{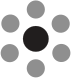}
    }
    $\xleftarrow{2}$
    \raisebox{0.8em}{
        \includegraphics[width=0.2in,angle=-90]{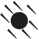}
    }
    $\xleftarrow{2}$
    \raisebox{1.0em}{
        \includegraphics[width=0.2in,angle=-90]{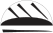}
    }

    \caption{Relative values of digits in the main proto-Elamite number systems. $X \xleftarrow{n} Y$ means that one $X$ has the same value as $n$ $Y$s. }
    \label{fig:systems}
\end{figure}

Proto-Elamite (PE) is a partially-deciphered script unearthed at early 3rd millennium BCE sites across the Iranian plateau. 
This script was exclusively used to record spreadsheet-style administrative accounts, and well over half of the attested glyphs are known to be digits.
Despite this abundance of numeric notations, no large-scale quantitative analysis of PE numerals has ever been undertaken, and most prior work has focused on the adjoining \textit{text}.
This is likely because the script employs multiple distinct number systems, which use partially-overlapping sets of digits and occasionally assign distinct values to identical-looking sign shapes (Figure~\ref{fig:systems}).
Many numerals can be read according to two or more of these systems, and represent different values depending on the system used.

Some PE signs have been given tentative readings in prior work, and on the basis of these readings there appears to be a regular relationship between the kind of object recorded and the number system used to count it (in a manner not entirely dissimilar to the measure words found across East Asian languages). 
Knowing which system is in use for a given numeral therefore increases the possibility of understanding what category of object is recorded in the adjoining text, and thus opens new avenues for the ongoing decipherment of this script.

In this work, we consider the task of disambiguating which systems are used in ambiguous PE numeral notations in order that the values of these numerals may be determined.
We describe a simple rule-based technique to extract lists of possible readings from PE numeral notations, which allows us to give the first large-scale survey of PE numerals since~\citealt{Friberg1978} (whose manual analysis occurred at a time when fewer texts were known).
We then propose two disambiguation techniques, one based on the subset-sum problem and another which uses a bootstrap classifier~\cite{yarowsky-1995-unsupervised}. 
We describe the construction of a test set for evaluating PE numeral disambiguation models, and propose a novel approach to cautious rule selection which significantly improves the performance of a bootstrap classifier on our data. 
Our analysis shows how these techniques lead to a deeper understanding of this ancient and undeciphered writing system.

\section{Data \& Background}
\begin{figure}
    \centering
    \includegraphics[width=\columnwidth]{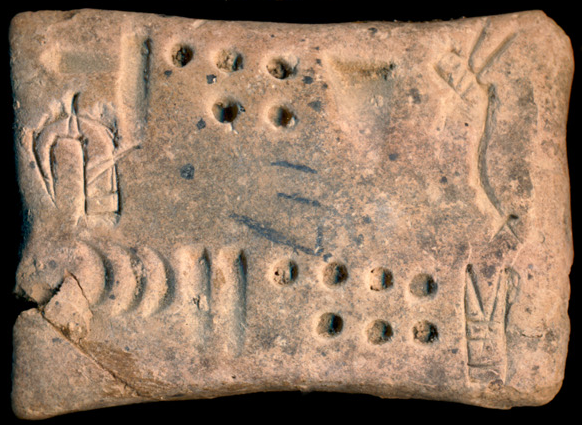}\\
    \begin{tabular}{ll}
       \textbf{Text} & \textbf{Numeral} \\
       \texttt{M056$\sim$f} & \texttt{1(N34) 5(N14) 1(N01) 1(N8B)} \\
       & $=111.5\times \texttt{N01}^S$\\
        \texttt{M341 M288} & \texttt{7(N14) 2(N01) 3(N39B)} \\
        & $=44.6\times\texttt{N01}^C$
    \end{tabular}
    \caption{Proto-Elamite tablet MDP 26, 177 (\citealt{mdp26}; P008805) alongside its transliteration and converted readings for both numerals. The tablet is read right-to-left: M056$\sim$f is the sign in the top-right corner. Observe that 111.5/44.6 equals the 2.5:1 ratio noted in Section~\ref{sec:mixed-systems}.}
    \label{fig:tablet-example}
\end{figure}

We base our analysis off the transliterated PE corpus hosted by the CDLI.\footnote{Cuneiform Digital Library Initiative, \url{https://cdli.mpiwg-berlin.mpg.de}; corpus downloaded 3 Oct 2022.} Each text in this corpus contains an optional header, followed by a series of ``entries'' written one per line of the transliterated file. Each entry contains a (possibly empty) span of text, a comma delimiter, and a (possibly empty) numeral notation. The transliterations use a work-in-progress signlist that reflects experts' current understanding of the texts, but which may not exactly match the true character inventory of the underlying script. In this signlist, characters that are believed to represent text are transliterated with labels beginning in ``M'' (e.g.\ M001, after \citeauthor{Meriggi1971la} who pioneered the study of this script), and those representing digits are labeled with ``N'' (e.g.\ N01). The notation \texttt{n(N00)} means that the digit \texttt{N00} is written down $n$ times. Figure~\ref{fig:tablet-example} shows an example of a tablet alongside its transliteration.

Most proto-Elamite numerals are written using one of four\footnote{Additionally, there are marginal systems (labeled B\#, C\#, and C'') which appear to be derived from one of the four main systems by the addition of hatch marks or boxes drawn around the digits. 
These systems are rare, and the extra hatching or boxing makes them trivial to identify, so we ignore them for the remainder of this work.
} number systems, which are called \textit{decimal} (D), \textit{sexagesimal} (S), \textit{bisexagesimal} (B), and \textit{capacity} (C).
In spite of their names, most of these systems use mixed radices. 
Figure~\ref{fig:systems} shows the relative values of the digits in each of these systems, as derived through prior manual analyses \cite{Friberg1978,DamerowEnglund1989,dahl2019}. 
Unlike Hindu-Arabic notation, where the value of a digit depends on its position, the values of proto-Elamite digits are fixed, and larger values are denoted by repeating a digit multiple times. 

Note that some digits can be used with distinct values in multiple systems (e.g.\ \texttt{N14} \NFourteen equals $10\times$\texttt{N01} \NOne in S, but only $6\times$\texttt{N01} \NOne in C): this means that it can sometimes be impossible to determine the absolute value of a numeral unless context makes clear which system it employs.

Since \texttt{N01} \NOne occurs as part of every number system, we use this sign as a standard unit and report values as multiples of \texttt{N01} whenever we convert to Hindu-Arabic notation. 
The S, D, and B systems are understood to represent unitless cardinal numbers, 
while C records unit\textit{ful} measures of volume. 
When it is necessary to emphasize that these systems are not commensurable, we add a superscript to denote the system in use: e.g.\ $12 \times \texttt{N01}^C$ is a measure of volume which is not equivalent to $12 \times \texttt{N01}^D$, despite having identical magnitude.

We summarize the above points with an illustrative example. The following notation (read from right-to-left)

\begin{center}
    \raisebox{0.4em}{\includegraphics[width=0.4in,angle=-90]{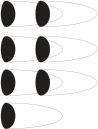}}
    \raisebox{0.4em}{\includegraphics[width=0.2in,angle=-90]{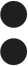}}
    \raisebox{0.8em}{\includegraphics[width=0.4in,angle=-90]{figures/signs/1N45.png}}
\end{center}

\noindent
would be transliterated as \texttt{1(N45) 2(N14) 7(N01)}. 
This numeral must use either the S or C system, as the large circle \texttt{N45} only occurs in these systems (Figure~\ref{fig:systems}).
Using the readings from the S system, this notation encodes a value of \\[-0.5em]

\noindent
\begin{tabular}{llll}
    $\hspace{-0.4em}3627\times \texttt{N01}\hspace{-1em}$ 
    &$= (1\times3600\hspace{-1em}$ 
    &$+ 2\times10\hspace{-1em}$ 
    &$+ 7\times1)\times \texttt{N01}\hspace{-1em}$ \\[0.5em]
    \multicolumn{4}{l}{\normalsize\hspace{-0.5em}Using the C system, it instead encodes}\\[0.5em]
    $\hspace{-0.4em}79\times \texttt{N01}\hspace{-1em}$ 
    &$= (1\times60\hspace{-1em}$ 
    &$+ 2\times6\hspace{-1em}$ 
    &$+ 7\times1)\times \texttt{N01}\hspace{-1em}$\\[0.5em]
\end{tabular}

\noindent
Of these, we know that the S reading must be the correct one, since \texttt{7(N01)} should never occur in the C system (every 6 \texttt{N01} would be bundled into an \texttt{N14}, so the actual notation for $79\times\texttt{N01}^C$ would be \texttt{1(N45) 3(N14) 1(N01)}).

It is not clear whether these notations would have been considered ambiguous at the time they were written. It is very plausible that the original scribes would have been able to infer the correct reading for a numeral based on contextual information which is not salient to the modern reader (for example, by knowing that certain items are consistently counted with a particular number system, a possibility discussed by~\citealt{englund2004,englund2011}). 
There is equally the possibility that these notations were ambiguous even to their authors, but that this ambiguity did not interfere with their intended use, for example if the PE texts were intended for short-term use when the scribes would still recall which system they intended at the time of writing.

\section{Methodology}
\subsection{Automated Conversion}
We extract all of the numeral notations from the transliterated corpus by using a regular expression to find every contiguous sequence of N-signs; we discard sequences which are damaged, which we identify as being immediately adjacent to a transliterated \texttt{X} or \texttt{...}. 
Algorithm~\ref{alg:conversion} uses the relative values from Figure~\ref{fig:systems} to automatically extract a dictionary of possible readings for each of these numerals. 

\begin{algorithm}[htbp]
\caption{PE Numeral Readings}\label{alg:conversion}
\begin{algorithmic}
\State \textbf{Input:} $digits = [(n_1, sign_1), ..., (n_k, sign_k)]$ 
\Comment{{\color{gray}A list of signs and num. times each one occurs.}}
\State \textbf{Returns:} A map from number systems to possible readings for this digit list.
\\\hrulefill
\For{$sys\in\{S, D, B, C\}$}
    \State $value_{sys} \gets 0$
\EndFor
\For{$(n, sign) \in digits$}
    \For{$sys\in\{S,D,B,C\}$}
        \If{$sign\notin signs\_used\_by(sys)$}
            \State $value_{sys} \gets \bot$
            \State$\triangleright\ ${\color{gray}$\bot$ means there is no valid reading in this system.}
            \State\textbf{continue}
        \EndIf
        \If{$n > max\_count(sign, sys)$}
            \State$\triangleright\ ${\color{gray}$max\_count$ returns the max num. of times this digit can occur before it would carry over to a higher value digit.}
            \State $value_{sys} \gets \bot$
        \EndIf
        \State $v \gets \mathrm{value\ of\ } sign \mathrm{\ in\ } sys$
        \State $value_{sys} \gets value_{sys} + n\times v$
        \State$\triangleright\ ${\color{gray}$\bot$ plus anything equals $\bot$}
    \EndFor
\EndFor
\State
\Return{$\{sys\!\mapsto\!value_{sys} \forall\!sys\in\{S,D,B,C\}\}$}
\end{algorithmic}
\end{algorithm}

Of the 8011 intact numerals which we have extracted, there are 7984 for which this conversion returns at least one reading. Of these, only 1899 unambiguously belong to a particular number system: the remainder are ambiguous between two, three, or even all four systems (Table~\ref{tab:initial-readings}). The following sections outline two proposals for disambiguating the ambiguous cases. Section~\ref{sec:failed-conversion} discusses the 27 cases for which there is no valid reading in any system.

\begin{table}[h]
    \centering
    \begin{tabular}{ll}
    \toprule
        \textbf{Possible Readings} & \textbf{Number of Numerals} \\\midrule
        none & 27 \\
        \midrule
        B & 18 \\
        C & 1678 \\
        D & 96 \\
        S & 107 \\
        \midrule
        B or D & 22 \\
        B or S & 5 \\
        C or D & 49 \\
        C or S & 143 \\
        \midrule
        B, C, or S & 185 \\
        B, D, or S & 292 \\
        \midrule
        B, C, D, or S & 5389 \\
        \bottomrule
    \end{tabular}
    \caption{Distribution of readings produced by our automated conversion. A majority of numerals in the corpus can be read using \textit{any one} of the four number systems.}
    \label{tab:initial-readings}
\end{table}

\subsection{Subset-Sum Analysis}\label{sec:subset-sum}
Our first approach to disambiguation relies on the fact that some PE documents end in a summary line, which records the total sum of the preceding entries. 
Although the entries themselves may be ambiguous, the sums naturally record larger amounts and are therefore more likely to use high-magnitude digits that unambiguously belong to a particular system.
When a tablet records values from multiple number systems, they are summarized separately; thus if we can identify unambiguous summaries, we can infer that all of the entries which they sum must belong to the same system.

To achieve this, we filter the corpus to find texts with one or two entries on the reverse, as current understandings of the corpus suggest that these are likely to be summaries.\footnote{Some transliterations include an annotation which explicitly labels a particular entry as a summary. However, not all summaries are labeled in this way, so we rely on automatic detection of summaries to expand the number of texts available for this analysis.}
For each of these texts, we solve an instance of the subset-sum problem to identify whether any combination of readings from the obverse adds up to the same value as any reading of the reverse. 

If an accurate summation is found, and any of the component terms has an unambiguous number system, we use this as evidence to disambiguate the entire text to that system. We manually evaluate this approach by confirming with domain experts whether the resulting disambiguations are correct.

\subsection{Bootstrapping}
Some of the PE numeral notations are inherently unambiguous, either because they use a digit which only occurs in a single system, or because they contain more instances of a digit than would be allowed by some systems.
We propose to use these cases as seed rules to train a bootstrap classifier~\cite{yarowsky-1995-unsupervised} for disambiguation. 

We choose bootstrapping because it requires only a small number of seed labels, and as seen in Table~\ref{tab:initial-readings} some systems have few unambiguous attestations. 
Moreover, bootstrapping yields interpretable results which can be understood by examining the label distribution associated with each input feature. 
This helps to legitimize model outputs to domain experts, and to situate model predictions relative to prior manual analyses.

\begin{table}[]
    \centering
    \scalebox{0.81}{
        \begin{tabular}{lp{2in}}
        \toprule
            \textbf{Feature} & \textbf{Description of Value(s)} \\
        \midrule        
        TABLET     & The tablet where this numeral occurs. \\
        FIRST\_SIGN & The first sign of the tablet where this numeral occurs (where we may expect to find a header). \\
        SAME\_ENTRY & Bag of signs which occur in the entry preceding this numeral. \\
        SAME\_TABLET & Bag of signs which occur anywhere on the same tablet as this numeral. \\
        OBJECT & The sign immediately preceding this numeral (where we may expect to find a counted object). \\
        IMPLICIT\_OBJECT & The last sign in the first entry of the text where this numeral occurs (where we may expect to find an implicit object). \\
        \bottomrule
        \end{tabular}
    }
    \caption{Each numeral is associated with a set of features from this list, which we use to train our bootstrap classifiers.}
    \label{tab:bootstrap-feats}
\end{table}

Table~\ref{tab:bootstrap-feats} lists the features used by our classifier. 
A numeral's initial label distribution is uniform over every system for which our automated conversion returns a valid reading, and zero elsewhere. 
We use the DL-2-ML algorithm~\cite{haffari-sarkar-analysis,DBLP:conf/acl/Abney02}, which models $\pi_x(j)$ (the likelihood that sample $x$ belongs to class $j$) as 

\[ \pi_x(j) \propto \prod_{f\in F_x}\theta_{fj} \]

\noindent
where $F_x$ is the set of features associated with sample $x$, and $\theta_{fj}$ is a learnable parameter which measures the association between feature $f$ and class $j$. 
We apply the ``cautious'' approach from~\citealt{collins-singer-1999-unsupervised}, which limits the number of rules that can be added to the decision list at each iteration of training. 
Specifically, candidate rules are sorted according to the number of labeled examples that support them, and only the $n$ with the largest support are added to the decision list.\footnote{\citealt{whitney-sarkar-2012-bootstrapping} note that many details are omitted from the description of the cautious algorithm in \citealt{collins-singer-1999-unsupervised}. We follow \citeauthor{whitney-sarkar-2012-bootstrapping} in assuming that confidence thresholding is performed using unsmoothed label counts. However, we differ from their approach by selecting the top $n$ rules overall (not the top $n$ for each label) as this yields stronger results on our data.}
$n$ starts at 5 and increases by 5 each iteration.

The cautious algorithm was motivated by the observation that ``the highest frequency rules [are] much `safer' [than low-frequency rules], as they tend to be very accurate''~\cite{collins-singer-1999-unsupervised}.
This observation does not seem to hold for our data, where many of the most frequent features barely meet the confidence threshold to be added to the decision list and, impressionistically, do not appear to be any more accurate than those with lower frequency.
We therefore propose a novel approach to cautious rule selection whereby $\theta_f$ is only updated if the update increases $\max_j \theta_{fj}$, i.e. if it increases the confidence of the label distribution associated with feature $f$.
In this setting we visit the features in a random order each iteration, to prevent a degenerate outcome whereby endless incremental updates are made only to the first rules visited.

\begin{table}[htb]
    \centering
    \begin{tabular}{ll}
    \toprule
        \textbf{System} & \textbf{Number of}  \\
                        & \textbf{Test Items} \\\midrule
        B & 3\\
        C & 18\\
        D & 14\\
        S & 13\\
        \bottomrule
    \end{tabular}
    \caption{Distribution of target classes in our numeral disambiguation test set. This set contains every instance of the B class which we were able to manually disambiguate with the help of domain experts; the other classes are kept small to maintain as balanced a distribution as possible.}
    \label{tab:test-distribution}
\end{table}

Prior to training, we upsample seeds with rare labels to obtain an equal number for each class.
We evaluate our classifiers on a test set which we construct by manually disambiguating some of the ambiguous notations in the corpus.
We endeavoured to keep this set as balanced as possible, but some systems (particularly B) can only be confidently identified in a few texts. 
This means the test set cannot contain every numeral for which we know the target label, as doing so would yield too great an imbalance between classes.
Appendix~\ref{sec:test-set} describes what evidence was used to disambiguate each numeral in the set, and Table~\ref{tab:test-distribution} summarizes the overall class distribution. 
All of the labels in the test set have been verified by domain experts.

\section{Results}
\subsection{Automated Conversion}
\subsubsection{Invalid Notations}\label{sec:failed-conversion}
There are 27 intact numerals for which our automated conversion does not return a valid reading according to any number system.
The vast majority of these violate the bundling principles established in prior work and shown in Figure~\ref{fig:systems}.
For example, the notation $\texttt{11(N01)}$ in P008043 should not be attested in any system, as every system carries over to a higher digit after at most ten \texttt{N01}s. 
These cases may be simple scribal errors, or they may suggest a lack of standardisation across scribes or across documents, which is consistent with the longstanding view that the writing system never achieved a significant degree of standardisation~\cite{dahl2019}.

More rarely, these conversion failures result from the presence of sign names such as \texttt{N02} which are not part of any PE number system established in prior work. 
Often these aberrant sign names correspond to digits from the related proto-cuneiform script. 
The number systems of proto-Elamite and proto-cuneiform bear a close resemblance, and in short tablets with minimal text content it can be difficult to confidently assign a text to one script or the other.
For these texts, expert evaluation may be needed to establish whether the aberrant sign names reflect simple transliteration errors or the erroneous inclusion of proto-cuneiform material in the proto-Elamite corpus.

\subsubsection{Mixed Systems}\label{sec:mixed-systems}
Our automated conversion reveals numerous accounts (Table~\ref{tab:mixed-systems}) which unambiguously use two different number systems (no tablets unambiguously use more than two systems).
In all but one of these, the C system occurs alongside one of the ``integer'' systems S, D, or B.
This suggests a general pattern of accounts which record capacities of goods received/disbursed from/to individual people, animals, households, or other entities also counted in whole numbers on the tablet.
The text P009383 is unique in that it unambiguously uses two of the ``integer'' systems S and D. 
On close inspection, however, the original tablet is heavily abraded where the putative S notation occurs, and the sign which forces this notation to be read as S (\texttt{N08}) is almost entirely unreadable. 
Given the otherwise total absence of texts which mix integer systems, we posit that this text may contain a transliteration error and that the broken sign is not in fact an \texttt{N08}.

\begin{table}[]
    \centering
    \begin{tabular}{ll}
    \toprule
        \textbf{Systems Used} & \textbf{Number of Tablets} \\

    \midrule
        C and S & 12 \\  
        C and D & 15 \\
        C and B & 4  \\
        S and D & 1 \\
    \bottomrule
    \end{tabular}
    \caption{Number of tablets which unambiguously use more than one number system.}
    \label{tab:mixed-systems}
\end{table}

Several texts which mix the S and C systems also have other features in common. 
P008796, P008798, and P008805 are exemplary of this group, which are all two-entry texts where the first entry is an S-denominated amount of M056$\sim$f 
\raisebox{1.4em}{\includegraphics[angle=-90,width=12pt]{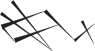}}
(possibly a plow), and the second is a C-denominated amount of M288. 
In all of these texts, there are exactly $2.5\times \texttt{N01}^S$ per $\texttt{N01}^C$: this ratio was previously identified and discussed by~\citet{DamerowEnglund1989,englund2004}.
P008791 appears to belong to this same class of texts, and given that the first entry records $128\times\texttt{N01}^S$ M056$\sim$f we should expect $51.2\times\texttt{N01}^C$ in the final entry (or \texttt{8(N14) 3(N01) 1(N39B)}).
We actually find $52\times\texttt{N01}^C$, or \texttt{8(N14) 4(N01)}. 
Close inspection of the tablet, however, reveals that there has been a transliteration error, and that the final sign, although mostly broken, is recognizably an \texttt{N39B}.
The text yields the expected ratio when this mistake is corrected.
Errors such as these are much easier to identify when dealing in converted Arabic numerals, which modern readers can understand and manipulate more quickly and intuitively than the original PE notations.

Out of 244 signs which precede at least two unambiguous notations, only 11 occur next to notations from distinct systems.
These 11 (M001, M056$\sim$f, M059, M096, M124, M218, M305, M327, M371, M387, M388) include signs which have been speculated~\cite{dahl2019} to represent human laborers or overseers (M388, M124), signs with possible syllabic values (M001, M096, M218, M387, M371), and headers or account owners (M059, M305, M327). 
In other words, these are signs which we expect to qualify or describe an object being counted, and not to be counted themselves.
It therefore appears that, while counted objects are consistently recorded using one particular number system, these qualifying signs can potentially be used to qualify objects from several different systems.
This suggests a novel approach to determine the function of signs with unknown meanings, by looking at the variety of number systems they occur beside.

\subsection{Subset-Sum Analysis}
Our subset-sum analysis identifies 24 texts which, upon manual inspection, can be fully or partially disambiguated based on their summary line(s). 
We highlight one which is of particular interest. In P008014, all entries must use the C system in order to equal the same value as the unambiguous C summary on the reverse. 
The text of the summary contains only the ``grain container'' sign M288, implying that the entire tablet should record amounts of M288. 
However, on the obverse of the tablet, M288 only occurs as the final sign of the very first entry.
This suggests that the scribe has only explicitly marked the counted object in the first entry, and left it implicit in the following entries (this is in keeping with known practices from other ancient Mesopotamian accounting corpora;~\citealt{nissen1993archaic}, pp. 37--38, \citealt{englund2001grain-accounting}).
This means that there exist long-distance dependencies between the entries in some texts, which need to be accounted for if these texts are to be fully understood.

\subsection{Bootstrapping}
\begin{figure}
    \centering
    \includegraphics[height=1.25in]{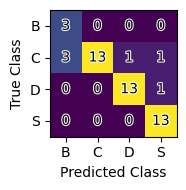}
    \includegraphics[height=1.25in]{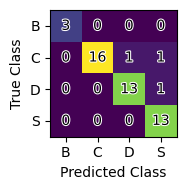}
    \caption{Confusion matrices from classifiers trained using the vanilla bootstrap algorithm (left) and our proposed variant (right).}
    \label{fig:bootstrap-results}
\end{figure}

Figure~\ref{fig:bootstrap-results} and Table~\ref{tab:bootstrap-results} compare results from our two approaches to bootstrapping. 
The baseline, vanilla bootstrapping algorithm achieves a reasonable F1 of 0.88.
Recall of the B, D, and S classes is strong, but the model performs more poorly on C notations, assigning these to the wrong system in $\sim28\%$ of cases.

\begin{table}[htp]
    \centering
    \scalebox{0.9}{
        \begin{tabular}{l | ccc | ccc}
        \toprule
            & \multicolumn{3}{c|}{\textbf{4-way}} & \multicolumn{3}{c}{\textbf{2-way}} \\
            \textbf{Model}   & \textbf{prec.} & \textbf{rec.} &\textbf{F1} & \textbf{prec.} & \textbf{rec.} &\textbf{F1} \\\midrule
            Baseline & 0.88 & 0.88 & 0.88 & 0.90 & 0.90 & 0.90 \\
            Ours & \textbf{0.94} & \textbf{0.94} & \textbf{0.94} & \textbf{0.96} & \textbf{0.96} & \textbf{0.96} 
            \\\bottomrule
        \end{tabular}
    }
    \caption{Numeral disambiguation results. In the 4-way setting, we seek to identify exactly which number system is in use for each numeral. In the 2-way setting, we only seek to distinguish C notations from everything else.}
    \label{tab:bootstrap-results}
\end{table}

By contrast, our proposed approach to cautious rule selection yields 0.94 F1, and completely eliminates the confusion between the C and B systems.
This suggests that frequency-based caution may not be the most appropriate rule-selection heuristic for all datasets, and that, in settings where bootstrap classifiers remain viable, it may be worthwhile to explore alternative approaches to cautious rule selection.
We also note that our approach converges to a solution extremely quickly, completing after just 25 iterations of rule updates, versus more than 200 iterations for the baseline model.

\begin{figure}
    \centering
    \includegraphics[height=1in]{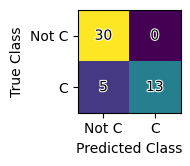}
    \includegraphics[height=1in]{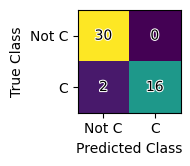}
    \caption{2-way disambiguation results. Confusion matrices from vanilla bootstrapping (left) and our proposed variant (right).}
    \label{fig:confusion-2way}
\end{figure}

Note from Figure~\ref{fig:systems} that ambiguities between the S, D, and B systems primarily come from the digits \texttt{N01} \NOne and \texttt{N14} \NFourteen, which have the same relative values across all three systems. S and B further overlap in the sign \texttt{N34} \NThirtyFour, which also maintains the same value across both systems. 
Thus many numerals which are technically ambiguous between these \textit{systems} nonetheless have unambiguous \textit{values}.\footnote{Here we assume that the \textit{absolute} value of \texttt{N01} is the same across all three systems, and not just its value relative to the other digits. Current understandings suggest that this is the case, but the undeciphered nature of the script means this is technically not certain.} 
In settings where the absolute value of a numeral is all that matters, it is therefore usually sufficient to distinguish these systems from C without distinguishing them from one another. 
In this two-way setting, our model achieves 0.96 F1, versus 0.90 F1 from the vanilla baseline (Figure~\ref{fig:confusion-2way}). 

We emphasize that our test set only contains numerals which domain experts were able to disambiguate based on manual inspection. 
Easy cases may therefore be over-represented, and we expect our results to be an upper bound on these classifiers' accuracy across the whole corpus.
Despite this, the results are strong enough to suggest that a majority of the ambiguous numerals in the corpus can be disambiguated with some certainty.

\section{Analysis}
In this section, we investigate how the features from our bootstrap classifier relate to known or hypothesized properties of the script.

A number of signs have been suggested as indicating types of livestock (sheep, goats, etc.); these include M346, M362, M367, and M417 \cite{dahl2005b}. In our analysis, all of these signs also predict the decimal system, which suggests that they were typically used to count flock sizes.

M376 has been suggested as either a "high-status human" \cite[p. 95]{dahl2005b} or livestock \cite[p. 165]{kelley2018}; it is strongly predicative of the sexigesimal system in our analysis.
Other objects associated with this system include M056$\sim$f (a plow), M219, M387$\sim$c and M320 (very speculatively, syllables used to write personal names), M059, M145, M153, and M365 (``owners'', possibly persons or institutions to whom these accounts belonged), and M269$\sim$c (a vessel?). 
This may cast some doubt on the livestock reading for M376, in favor of the high-status human reading which is a more natural fit among the owner signs and possible personal names in this collection.

We note one text, P009343,\footnote{An earlier version of this work mistakenly cited this text as P008212.} which may exhibit a consistent ratio related to M376.
P009343 alternates between entries ending in M288 and entries ending in M376 or M367$\sim$i. 
The magnitude of the numeral in an M288 entry is always exactly 4 times as large as the magnitude of an adjacent M376 entry, or 2 times as large as an adjacent M367$\sim$i entry.
This pattern is very unlikely to be due to chance, as it holds across 44 total entries. 
Whatever the meaning of M376, on the basis of this text we can assert that it is associated with amounts of M288 that are exactly twice as large as those associated with M367$\sim$i. 
To our knowledge this ratio has not yet been noted in previous publications.

After disambiguating every numeral in the corpus to the most likely system according to our bootstrap classifier, we measure the average magnitude of counts associated with each feature.
We observe that certain features accompany significantly higher or lower counts than others.
Entries ending in M288 have the largest capacity magnitudes on average, while those ending in M263 are among the smallest. 
Both signs have been speculated to represent containers;
from our results one might further speculate that M263 is a container of smaller dimension, or one that was never dealt with in bulk quantities.
Entries ending in M297$\sim$b also accompany unusually large capacity measures, though the visually-similar M297 does not stand out as unusually large or small.
This may point towards these signs having distinct meanings or uses despite sharing a visual resemblance. 

The numerical systems of proto-Elamite have been proposed to have functional uses relating to cultural practices in 3rd millennium south-western Iran. For instance, the capacity system (C) is suggested to be used for counting rations disbursed to households or workers \cite[pgs. 153-155]{kelley2018}.
Among the recipients of these rations are M388 and M124, parallel ``worker categories'' which may represent the heads of work teams. 
We find that tablets whose first entry ends in M388 contain significantly larger capacity measures on average than those ending M124, which may point towards M388 individuals heading teams of larger sizes or comprising workers of higher status.

\section{Related Work}
\citet{naik-etal-2019-exploring} demonstrate that word embedding models fail to capture magnitude and numeration (i.e. the equivalence between \textit{3} and \textit{three}), and suggest the need for dedicated representations of numerals in NLP models. 
\citet{sundararaman-etal-2020-methods} follow up with DICE embeddings designed to explicitly capture both magnitude and numeration, and demonstrate improved results on numeracy tests introduced by \citet{wallace-etal-2019-nlp}. 
\citet{spithourakis-riedel-2018-numeracy} describe a GMM-based approach to numeral embedding for language models, which also incorporates explicit representations of magnitude.
These models assume that the magnitudes in question are known and must simply be encoded; they do not consider the task of \textit{determining} magnitude from ambiguous notations.

While introducing a benchmark to test LM numeracy, \citet{shi-etal-2022-chainofthought} note that numeral representations can vary across scripts; however, they assume a setting where the conversion to Hindu-Arabic notation is straightforward, and do not discuss ambiguities which may result from this conversion.

One approach to handling numeric values in word problems is to replace them with variables $v_1, v_2, ...$, generate the solution as an equation in terms of these variables, and substitute the original values back to obtain a concrete solution.
\citet{wu-etal-2021-math} note that the choice of equation can sometimes depend on whether the original quantities were absolute values or percentages, and therefore this replacement can introduce ambiguities which make some problems unsolvable.
They introduce a magnitude-aware encoding for digit sequences, and describe a numerical properties prediction mechanism which estimates whether a numeral is an integer, fraction, percentage, etc.
This mirrors our attempt to predict an underlying number system.

\citet{berg-kirkpatrick-spokoyny-2020-empirical} investigate the task of predicting a numeric value given surrounding text as context. They find that models which implicitly separate a value's mantissa from its exponent achieve better results than those which predict the value directly, and that context from large pretrained text encoders is useful even when the pretraining task was not focused on promoting numeracy.
As we are dealing with an undeciphered corpus, our models are unfortunately unable to rely on pretrained embeddings for context.

\citet{Friberg1978} is responsible for early analyses of proto-Elamite and proto-cuneiform (a related script which is also partially deciphered) which helped to establish the relative values of the digits in these corpora.
\citet{NissenDamerowEnglund1994} perform what is possibly the earliest computer-assisted analysis of bookkeeping practices in proto-cuneiform, while
\citet{DamerowEnglund1989} and \citet{englund2011} discuss accounting practices in proto-cuneiform and proto-Elamite and the relationships between the two.

\section{Conclusion}
We have automated the process of extracting candidate readings for ancient proto-Elamite numeral notations, and have described ambiguities in the original script which make this extraction challenging.
We present two approaches for disambiguating these ambiguous notations: one exploits a common structural property of proto-Elamite accounts to look for unambiguous summations, and the other exploits the few unambiguous notations to train a bootstrap classifier.
We create a test set for the disambiguation task by manually disambiguating a subset of the corpus, and we describe a novel variant of cautious rule selection which significantly improves bootstrapping performance on this test set. 
As a result of this work, we are able to assign confident values to a majority of the numeral notations in proto-Elamite, to identify and correct a number of transliteration errors in the proto-Elamite corpus, and to shed new light on existing hypotheses about the meanings of some signs in this partially-deciphered script. 
As the proto-Elamite script was fundamentally an accounting technology, we believe that this work represents a crucial step towards deepening our understanding of this ancient corpus.

\section*{Limitations}
As PE remains largely undeciphered, our results can only be evaluated on the small subset of numerals which we have manually disambiguated.
As noted in the paper, these may be easier than the rest of the corpus, meaning our evaluation can only give an upper bound on model performance.

We use a feature-based classifier to give interpretable results which can more easily be shared and discussed with non-technical experts in Assyriology. A limitation of this approach is that model performance depends on the choice of input features, and features which are effective can sometimes seem arbitrary.
We attempt to justify the features used by our model by explaining in  Table~\ref{tab:bootstrap-feats} which aspects of the script each is intended to capture.

Lastly, PE numerals are just one aspect of a complex and multifarious decipherment problem. Our results alone cannot paint a complete picture of this script, and must be interpreted in relation to results from outside of computer science.

\bibliography{anthology,custom,pe-bib}

\begin{thebibliography}{25}
\expandafter\ifx\csname natexlab\endcsname\relax\def\natexlab#1{#1}\fi

\bibitem[{Abney(2002)}]{DBLP:conf/acl/Abney02}
Steven~P. Abney. 2002.
\newblock \href {https://doi.org/10.3115/1073083.1073143} {Bootstrapping}.
\newblock In \emph{Proceedings of the 40th Annual Meeting of the Association for Computational Linguistics, July 6-12, 2002, Philadelphia, PA, {USA}}, pages 360--367. {ACL}.

\bibitem[{Berg-Kirkpatrick and Spokoyny(2020)}]{berg-kirkpatrick-spokoyny-2020-empirical}
Taylor Berg-Kirkpatrick and Daniel Spokoyny. 2020.
\newblock \href {https://doi.org/10.18653/v1/2020.emnlp-main.385} {An empirical investigation of contextualized number prediction}.
\newblock In \emph{Proceedings of the 2020 Conference on Empirical Methods in Natural Language Processing (EMNLP)}, pages 4754--4764, Online. Association for Computational Linguistics.

\bibitem[{Collins and Singer(1999)}]{collins-singer-1999-unsupervised}
Michael Collins and Yoram Singer. 1999.
\newblock \href {https://aclanthology.org/W99-0613} {Unsupervised models for named entity classification}.
\newblock In \emph{1999 Joint {SIGDAT} Conference on Empirical Methods in Natural Language Processing and Very Large Corpora}.

\bibitem[{Dahl(2005)}]{dahl2005b}
Jacob~L. Dahl. 2005.
\newblock Animal husbandry in {S}usa during the proto-{E}lamite period.
\newblock \emph{Studi Micenei ed Egeo-Anatolici}, 47:81--134.

\bibitem[{Dahl(2019)}]{dahl2019}
Jacob~L. Dahl. 2019.
\newblock \emph{Tablettes et fragments Proto-{\'e}lamites / Proto-{E}lamite Tablets and Fragments}.
\newblock D\'epartment des Antiquit\'es Orientales.

\bibitem[{Damerow and Englund(1989)}]{DamerowEnglund1989}
Peter Damerow and Robert~K. Englund. 1989.
\newblock \emph{The proto-Elamite texts from Tepe Yahya}, volume~39 of \emph{American School of Prehistoric Research: Bulletin}.
\newblock Peabody Museum, Cambridge, Massachusetts.

\bibitem[{Englund(1996)}]{englund1996}
Robert~K. Englund. 1996.
\newblock The proto-{E}lamite script.
\newblock \emph{{The World's Writing Systems}}, pages 160--164.

\bibitem[{Englund(2001)}]{englund2001grain-accounting}
Robert~K. Englund. 2001.
\newblock Grain accounting practices in archaic {M}esopotamia.
\newblock In J.~H{\o}yrup and Peter Damerow, editors, \emph{Changing Views on Ancient Near Eastern Mathematics}, pages 1--35.

\bibitem[{Englund(2004)}]{englund2004}
Robert~K. Englund. 2004.
\newblock \href {https://cdli.ucla.edu/staff/englund/publications/englund2004c.pdf} {The state of decipherment of proto-{E}lamite}.
\newblock \emph{The First Writing: Script Invention as History and Process}, pages 100--149.

\bibitem[{Englund(2011)}]{englund2011}
Robert~K. Englund. 2011.
\newblock Accounting in proto-cuneiform.
\newblock In K.~Radner and E.~Robson, editors, \emph{The Oxford Handbook of Cuneiform Culture}, pages 32--50.

\bibitem[{Friberg(1978)}]{Friberg1978}
J\"oran Friberg. 1978.
\newblock \emph{The Third Millennium Roots of Babylonian Mathematics I-II}.
\newblock G\"oteborg Dept. of Mathematics, Chalmers University of Technology, G\"oteborg, Sweden.

\bibitem[{Haffari and Sarkar(2007)}]{haffari-sarkar-analysis}
Gholamreza Haffari and Anoop Sarkar. 2007.
\newblock \href {https://dl.acm.org/doi/proceedings/10.5555/3020488} {Analysis of semi-supervised learning with the {Y}arowsky algorithm}.
\newblock In \emph{Uncertainty in Artificial Intelligence (UAI 2007)}, pages 159 -- 166. AUAI Press.
\newblock Conference in Uncertainty in Artificial Intelligence 2007, UAI 2007 ; Conference date: 19-07-2007 Through 22-07-2007.

\bibitem[{Kelley(2018)}]{kelley2018}
Kathryn Kelley. 2018.
\newblock \href {https://ora.ox.ac.uk/objects/uuid:afa3362e-1182-43aa-a2b9-d675bd8c585a} {\emph{Gender, Age, and Labour Organization in the Earliest Texts from Mesopotamia and Iran (c. 3300--2900 BC).}}
\newblock Doctoral dissertation, University of Oxford.

\bibitem[{Meriggi(1971)}]{Meriggi1971la}
Piero Meriggi. 1971.
\newblock \emph{La scrittura proto-elamica. Parte Ia: La scrittura e il contenuto dei testi}.
\newblock Accademia Nazionale dei Lincei, Rome.

\bibitem[{Naik et~al.(2019)Naik, Ravichander, Rose, and Hovy}]{naik-etal-2019-exploring}
Aakanksha Naik, Abhilasha Ravichander, Carolyn Rose, and Eduard Hovy. 2019.
\newblock \href {https://doi.org/10.18653/v1/P19-1329} {Exploring numeracy in word embeddings}.
\newblock In \emph{Proceedings of the 57th Annual Meeting of the Association for Computational Linguistics}, pages 3374--3380, Florence, Italy. Association for Computational Linguistics.

\bibitem[{Nissen et~al.(1994)Nissen, Damerow, and Englund}]{NissenDamerowEnglund1994}
Hans~J. Nissen, Peter Damerow, and Robert~K. Englund. 1994.
\newblock \emph{Archaic Bookkeeping: Writing and Techniques of Economic Administration in the Ancient Near East}.
\newblock University of Chicago Press.

\bibitem[{Nissen et~al.(1993)Nissen, Damerow, and Englund}]{nissen1993archaic}
H.J. Nissen, P.~Damerow, and R.K. Englund. 1993.
\newblock \href {https://books.google.ca/books?id=YBAzXV4YtQ8C} {\emph{Archaic Bookkeeping: Early Writing and Techniques of Economic Administration in the Ancient Near East}}.
\newblock University of Chicago Press.

\bibitem[{Scheil(1935)}]{mdp26}
Vincent Scheil. 1935.
\newblock \emph{Texte de comptabilit\'{e} Proto-\'{E}lamite ({T}roisi\`{e}me S\'{e}rie)}, volume~26 of \emph{M\'{e}moires de la Mission Archa\'{e}ologique de Perse}.
\newblock Paris: Librairie Ernest Leroux.

\bibitem[{Shi et~al.(2022)Shi, Suzgun, Freitag, Wang, Srivats, Vosoughi, Chung, Tay, Ruder, Zhou, Das, and Wei}]{shi-etal-2022-chainofthought}
Freda Shi, Mirac Suzgun, Markus Freitag, Xuezhi Wang, Suraj Srivats, Soroush Vosoughi, Hyung~Won Chung, Yi~Tay, Sebastian Ruder, Denny Zhou, Dipanjan Das, and Jason Wei. 2022.
\newblock \href {https://doi.org/10.48550/arXiv.2210.03057} {Language models are multilingual chain-of-thought reasoners}.
\newblock \emph{CoRR}, abs/2210.03057.

\bibitem[{Spithourakis and Riedel(2018)}]{spithourakis-riedel-2018-numeracy}
Georgios Spithourakis and Sebastian Riedel. 2018.
\newblock \href {https://doi.org/10.18653/v1/P18-1196} {Numeracy for language models: Evaluating and improving their ability to predict numbers}.
\newblock In \emph{Proceedings of the 56th Annual Meeting of the Association for Computational Linguistics (Volume 1: Long Papers)}, pages 2104--2115, Melbourne, Australia. Association for Computational Linguistics.

\bibitem[{Sundararaman et~al.(2020)Sundararaman, Si, Subramanian, Wang, Hazarika, and Carin}]{sundararaman-etal-2020-methods}
Dhanasekar Sundararaman, Shijing Si, Vivek Subramanian, Guoyin Wang, Devamanyu Hazarika, and Lawrence Carin. 2020.
\newblock \href {https://doi.org/10.18653/v1/2020.emnlp-main.384} {Methods for numeracy-preserving word embeddings}.
\newblock In \emph{Proceedings of the 2020 Conference on Empirical Methods in Natural Language Processing (EMNLP)}, pages 4742--4753, Online. Association for Computational Linguistics.

\bibitem[{Wallace et~al.(2019)Wallace, Wang, Li, Singh, and Gardner}]{wallace-etal-2019-nlp}
Eric Wallace, Yizhong Wang, Sujian Li, Sameer Singh, and Matt Gardner. 2019.
\newblock \href {https://doi.org/10.18653/v1/D19-1534} {Do {NLP} models know numbers? probing numeracy in embeddings}.
\newblock In \emph{Proceedings of the 2019 Conference on Empirical Methods in Natural Language Processing and the 9th International Joint Conference on Natural Language Processing (EMNLP-IJCNLP)}, pages 5307--5315, Hong Kong, China. Association for Computational Linguistics.

\bibitem[{Whitney and Sarkar(2012)}]{whitney-sarkar-2012-bootstrapping}
Max Whitney and Anoop Sarkar. 2012.
\newblock \href {https://aclanthology.org/P12-1065} {Bootstrapping via graph propagation}.
\newblock In \emph{Proceedings of the 50th Annual Meeting of the Association for Computational Linguistics (Volume 1: Long Papers)}, pages 620--628, Jeju Island, Korea. Association for Computational Linguistics.

\bibitem[{Wu et~al.(2021)Wu, Zhang, Wei, and Huang}]{wu-etal-2021-math}
Qinzhuo Wu, Qi~Zhang, Zhongyu Wei, and Xuanjing Huang. 2021.
\newblock \href {https://doi.org/10.18653/v1/2021.acl-long.455} {Math word problem solving with explicit numerical values}.
\newblock In \emph{Proceedings of the 59th Annual Meeting of the Association for Computational Linguistics and the 11th International Joint Conference on Natural Language Processing (Volume 1: Long Papers)}, pages 5859--5869, Online. Association for Computational Linguistics.

\bibitem[{Yarowsky(1995)}]{yarowsky-1995-unsupervised}
David Yarowsky. 1995.
\newblock \href {https://doi.org/10.3115/981658.981684} {Unsupervised word sense disambiguation rivaling supervised methods}.
\newblock In \emph{33rd Annual Meeting of the Association for Computational Linguistics}, pages 189--196, Cambridge, Massachusetts, USA. Association for Computational Linguistics.

\end{thebibliography}
\bibliographystyle{acl_natbib}

\appendix

\newpage
\section{Constructing a Test Set for Numeral Disambiguation}\label{sec:test-set}

This appendix describes a set of ambiguous numerals which we believe can be confidently disambiguated. Together, these numerals comprise the test set which we use to evaluate automated disambiguation via bootstrapping.

\subsection{Capacity Measures}\label{sec:capacity-analysis}
\paragraph{P008014}
Our subset-sum analysis (Section~\ref{sec:subset-sum}) reveals that the sole entry on the reverse of this tablet (P008014:18:num) exactly equals the sum of the entries on the obverse, provided the whole tablet is read in the capacity system. 
This is consistent with the fact that four of the entries on the text are unambiguously capacity measures (P008014:11:num, P008014:14:num, P008014:15:num, and P008014:18:num; the other entries are SDBC-ambiguous).
Moreover, the apparent summary line has an M288 object, and the first entry on the obverse ends in M288. This suggests that the entire text may record amounts of M288 (which is strongly associated with the capacity system), but that this object has been left implicit in all but the first and last entries.

On the basis of this evidence, we disambiguate the seven ambiguous entries in this text (P008014:6:num, P008014:7:num, P008014:8:num, P008014:9:num, P008014:10:num, P008014:12:num, P008014:15:num) to the capacity system.

\subsection{Sexagesimal Measures}
\paragraph{P008173}
Our subset-sum analysis reveals that the first entry on the reverse (P008173:13:num), which is an unambiguous sexagesimal notation counting $7.5 \times$ \texttt{N01$^S$} instances of M376, exactly equals the sum of the M376 entries on the obverse (P008173:5:num, P008173:7, P008173:8) if these entries are read in the sexagesimal system. On the basis of this evidence we disambiguate these entries to the sexagesimal system.

\paragraph{M056$\sim$f/M288 Texts}
P008798 is exemplary of a set of two-entry texts of comparable physical dimensions (approx. 43mm x 31mm x 18mm) which count M056$\sim$f in the first entry and M288 in the second. In many of these texts, the first entry is unambiguously sexagesimal and the second is unambiguously capacity; in these cases the amount of M056$\sim$f is always exactly 2.5 times the amount of M288.

If P008797:6:num is read as a sexagesimal notation, then this text follows the same pattern as these other texts, and exhibits the expected 2.5:1 ratio of M056$\sim$f to M288. On the basis of this evidence we disambiguate this entry to the sexagesimal system.

By the same argument, we also disambiguate P008791:6, P008799:6, P008800:6, P008801:6, P008802:6, P008804:4 and P008810:7 to the sexagesimal system.

\subsection{Decimal Measures}
\paragraph{P008179}
This tablet contains an unambiguous summary entry ending in M388 with a value of $1412\times$ \texttt{N01$^D$}. 
This is exactly the value of the entries on the obverse (which also end in M388) when read in the decimal system. 

\paragraph{P008012}
Following the claim that sheep and goats are counted decimally in proto-Elamite~\cite{englund2011}, the entirety of this text should be expected to use the decimal system. Our subset-sum analysis confirms that the entry on the reverse (P008012:16) equals the sum of the entries on the obverse when read in this system, though in this case that would also be true for sexagesimal and bisexagesimal readings. Notably, the summation does \textit{not} work out if the summary is read as a capacity measure. Thus, while we may confidently say that this text does not contain capacity measures, we can only tentatively assign it to the decimal system in particular.

\paragraph{P008243}
\citet{kelley2018} observes that this is a decimally-counted roster. We follow this author in disambiguating all ambiguous numerals in this text to the decimal system.

\subsection{Bisexagesimal Measures}
\paragraph{P009048}
This text contains a large number of unambiguous bisexagesimal notations. Of these, P009048:16:num counts the same object (M352$\sim$h) as P009048:19, on the basis of which we propose that P009048:19 is also a bisexagesimal notation. Similarly, the unambiguous entries P009048:10 and P009048:15 count the same objects (M351+X and M354, respectively) as P009048:13 and P009048:7, respectively, on the basis of which we also disambiguate these entries to B.

\end{document}